\documentclass[12pt]{article}
\DeclareRobustCommand*\cal{\relax\mathcal}
\usepackage{fullpage}
\usepackage{amssymb, amsthm, amsmath}
\usepackage{doublespace}
\usepackage{bm}
\usepackage{graphicx}
\usepackage[authoryear]{natbib}
\usepackage{bm}
\usepackage{verbatim}
\usepackage{lineno}
\usepackage{times}
\usepackage{tabulary}
\usepackage{url}
\usepackage{gensymb}
\usepackage{subcaption}
\usepackage{booktabs}
\usepackage{authblk}

\newcommand{\beq}{ \begin{equation}}
\newcommand{\eeq}{ \end{equation}}
\newcommand{\beqn}{ \begin{eqnarray}}
\newcommand{\eeqn}{ \end{eqnarray}}

\title{Nearest-Neighbor Neural Networks for Geostatistics}
\author[1]{Haoyu Wang}
\author[1]{Yawen Guan}
\author[1]{Brian J Reich}
\affil[1]{Department of Statistics, North Carolina State University}

\begin{document}
\maketitle

\begin{abstract}\begin{singlespace}
\noindent Kriging is the predominant method used for spatial prediction, but relies on the assumption that predictions are linear combinations of the observations.  Kriging often also relies on additional assumptions such as normality and stationarity.  We propose a more flexible spatial prediction method based on the Nearest-Neighbor Neural Network (4N) process that embeds deep learning into a geostatistical model. We show that the 4N process is a valid stochastic process and propose a series of new ways to construct features to be used as inputs to the deep learning model based on neighboring information. Our model framework  outperforms some existing state-of-art geostatistical modelling methods for simulated non-Gaussian data and is applied to a massive forestry dataset. \vspace{12pt}
\end{singlespace}\end{abstract}
\newpage
The Gaussian process (GP) \citep{rasmussen2004gaussian} is the foundational stochastic process used in geostatistics \citep{cressie1992statistics, stein2012interpolation, gelfand2010handbook, gelfand2016spatial}. GPs are used directly to model Gaussian data and as the basis of non-Gaussian models such as generalized linear \citep[e.g.,][]{diggle1998model}, quantile regression \citep[e.g.,][]{lum2012spatial,reich2012spatiotemporal} and spatial extremes \citep[e.g.,][]{cooley2007bayesian,sang2010continuous} models. Similarly, Kriging is the standard method for geostatistical prediction. Kriging can be motivated as the prediction that arises from the conditional distribution of a GP or more generally as the best linear unbiased predictor under squared error loss \citep{ferguson2014mathematical, whittle1954stationary, ripley2005spatial}. Both GP modeling and Kriging require estimating the spatial covariance function, which often relies on assumptions such as stationarity.  Parametric assumptions such as linearity, normality and stationary can be questionable and difficult to verify. 

Flexible methods have been developed to overcome these limitations \citep[provide a review]{reich2015spatial}. There is an extensive literature on nonstationary covariance modelling \citep[provides a recent review]{risser2016nonstationary}. Another class of methods estimates the covariance function nonparametrically but retains the normality assumption \citep[e.g.,][]{huang2011nonparametric,im2007semiparametric,choi2013nonparametric}. Fully nonparametric methods that go beyond normality have also been proposed \citep[e.g.,][]{gelfand2005bayesian,duan2007generalized,reich2007multivariate,rodriguez2011nonparametric}, but are computationally intensive and often require replication of the spatial process.

Deep learning has emerged as a powerful alternative framework for prediction.  The ability of deep learning to reveal intricate structures in high-dimensional data such as images, text and videos make them extremely successful for complex tasks \citep[][provide detailed reviews]{schmidhuber2015deep, goodfellow2016deep, lecun2015deep}. For similar reasons, researchers have begun to explore the application of deep learning to model complex spatial-temporal data. \cite{xingjian2015convolutional}  propose the convolutional long short term memory (ConvLSTM) structure to capture spatial and temporal dependencies in precipitation data. \cite{zhang2017deep} design an end-to-end structure of ST-ResNet based on original deep residual learning \citep{he2016deep} for citywide crowd flows prediction. \cite{fouladgar2017scalable} propose a convolutional and recurrent neural network based method to accurately predict traffic congestion state in real-time based on the congestion state of the neighboring information. \cite{Rodrigues2018BeyondED} propose a multi-output multi-quantile deep learning approach for jointly modeling several conditional quantiles together with the conditional expectation for spatial-temporal problems. One common structure these previous work apply is convolutional neural network, which requires input data presented in grid cell form, i.e., an image. However, less attention in this field has been drawn to point-referenced geostatistical data at irregular locations (non-gridded). \cite{di2016assessing} establish a hybrid geostatistical model that incorporates multiple covariates (including a convolutional layer to capture neighboring information) to improve model performance. Unlike the proposed method, neighboring observations enter the predictive model of \cite{di2016assessing} only through a linear combination (similar to the 4N-Kriging model described below in Section 2.3).

In this paper, we establish a more flexible spatial prediction method by embedding deep learning into a geostatistical model.  The method uses deep learning to build a predictive model based on neighboring values and their spatial configuration.  We show that our Nearest-Neighbor Neural Network (4N) is a valid stochastic process model that does not rely on assumptions of stationarity, linearity or normality.  Different sets of features are created as inputs to the 4N model including (but are not restricted to) Kriging predictions, neighboring information and spatial locations.  By including different features, 4N spans methods that are anchored on parametric models to methods that are completely nonparametric and make no assumptions about the relationship between nearby observations.  By construction, the flexible method can be implemented using standard and highly-optimized deep learning software and can thus be applied to massive datasets.

The remainder of the paper is organized as follows. Section 2 presents the 4N model. Section 3 compares the proposed model with the Nearest-Neighbor Gaussian Process model \citep{datta2016} and the spatial Asymmetric Laplace Process model \citep{lum2012spatial} via a simulation study. In Section 4, the algorithm is applied to Canopy Height Model (CHM) data. Section 5 summarizes the results and discusses a potential method to obtain prediction uncertainties.

\section{Methodology}
\subsection{Nearest-Neighbor Gaussian Process}

In this section we review the Nearest-Neighbor Gaussian Process (NNGP)  of \cite{datta2016}.  Let $Y_i$ be the response at spatial location $s_i$, and denote $Y_{\cal N} = \{Y_i;\ i \in \cal N\}$ for any set of indices $\cal N$. The NNGP is defined in domain $\cal D $ and is specified in terms of a reference set $\cal S=\{1,...,n\}$, which we take to be observation locations, and locations outside of $\cal S$, $\cal U=\{n+1,...,n+k\}$. The joint density of $Y_{\cal S}$ can be written as the product of conditional densities,
\begin{equation}
p(Y_{\cal S}) = \prod_{i=1}^n p(Y_i|Y_1,...,Y_{i-1}).
\end{equation}
When $n$ is large this joint distribution is unwieldy and \cite{datta2016} use the conditioning set approximation \citep{vecchia1988,stein2004,gramacy2015,gramacy2014, datta2016}
\begin{equation}\label{e:NNGP2}   p(Y_{\cal S})   \approx \tilde{p}(Y_{\cal S}) = 
\prod_{i=1}^n p(Y_{i}|Y_{\cal N_i})
\end{equation}
where $\cal N_i\subset\{1,...,i-1\}$ is the conditioning set (also referred to as the neighboring set) for observation $i$ of size $m << n$.   \cite{datta2016} and \cite{lauritzen1996} prove that  $\tilde{p}(Y_{\cal S})$ constructed in this way is a proper joint density.

We follow \cite{vecchia1988}'s choice of neighboring set and take $\cal N_i$ to be the indices of the $m$ nearest neighbors of $s_i$ in the set $\{s_1, ..., s_{i-1}\}$. The motivation for this approximation is that after conditioning on the $m$ nearest (and thus most strongly correlated) observations, the remaining observations do not contribute substantially to the conditional distribution and thus the approximation is accurate. This approximation is fast because it only requires dealing with matrices with size of $m\times m$.  

\cite{datta2016} further define the conditional density of $Y_{\cal U}$ as 
\begin{equation}
p(\mathbf{Y_{\cal U} | Y_{\cal S}}) = \prod_{i=n+1}^{n+k} p(Y_{i}| Y_{\cal N_i}),
\end{equation}
where $\cal N_{i}$ for $i>n$ is comprised of the $m$ nearest neighbors to $s_i$ in $\cal S$. This is a proper conditional density. Equations (2) and (3) are sufficient to describe the joint density of any finite set over domain $\cal D$. \cite{datta2016} show that these densities satisfy the Kolmogorov consistency criteria and thus correspond to a valid stochastic process.

\subsection{Nearest-Neighbor Neural Network (4N) model}

While the sparsity of the NNGP delivers massive computational benefits, it remains a parametric Gaussian model. We note that the proof in \cite{datta2016} that the NNGP model for $(Y_{\cal S}, Y_{\cal U})$ yields a valid stochastic process holds for any conditional distribution $p(Y_i| Y_{\cal N_i})$, including non-linear and non-Gaussian conditional distributions. Therefore, in this paper we extend the nearest neighbor process to include a neural network in the conditional distribution and thus achieve a more flexible modeling framework. 

Let $\mathbf{X_i} = (X_{i1},...,X_{ip})$ be features constructed from the available information $s_i$, $\cal N_i$ and $Y_{\cal N_i}$ (this can include spatial covariates such as the elevation corresponding to $s_i$).  The features are then related to the response via the link function $f(\mathbf{X_i})$, that is, $Y_i = f(\mathbf{X_i}) + \epsilon_i$, where $\epsilon_i$ is additive error.  The NNGP model assumes that $f(\cdot)$ is a linear combination of $Y_{\cal N_i}$ with weights determined by the configuration of $s_i$ and locations in its neighboring set.  We generalize this model using a multilayer perceptron for $f(\mathbf{X}_i)$.  A multilayer perceptron (MLP) is a class of feedforward neural network. It consists of at least three layers of nodes: an input layer, one or more than one hidden layers, and an output layer. Nodes in different layers are connected with activation function such as Relu, tanh and sigmoid. The hidden layers and nonlinear activation function enable multilayer perceptron capture nonlinear relationships.

Let $\mathbf{X} = (\mathbf{X_1^T}, \mathbf{X_2^T}, ..., \mathbf{X_n^T})^T$ be the $n \times p$ covariate matrix, where $\mathbf{X_i}, i=1,2,...,n$ is associated with corresponding $Y_{\cal N_i}$ (we will explain how to generate these features in the next section). Our 4N model is then:
\begin{equation}
    Y_i = f(\mathbf{X_i}) + \epsilon_i,
\end{equation}
where $f(\cdot)$ is a multilayer perceptron and $\epsilon_i$ is iid random error. In matrix form, an $L$-layer perceptron can be written as follows:
\begin{align*} 
\mathbf{Z^1} &= \mathbf{W^1X^T} + \mathbf{b^1}, \;\;\;  \mathbf{A^1} = \psi_1(\mathbf{Z^1}) \\ 
\mathbf{Z^2} &= \mathbf{W^2A^1} + \mathbf{b^2}, \;\;\;   \mathbf{A^2} = \psi_2(\mathbf{Z^2}) \\
... \\
\mathbf{Z^L} &= \mathbf{W^L A^{L-1}} + \mathbf{b^L},  \;\;\;  f(\mathbf{X}) = \psi_L (\mathbf{Z^L}),
\end{align*}
where $\psi_l(\cdot)$ is an activation function for layer $l$ and $f(\mathbf{X}_i)$ is the $i^{th}$ element of $f(\mathbf{X})$. The activation function for the $L$th layer, also known as the output layer, is chosen according to the problems at hand. For instance, in this paper, we focus on regression with real-valued responses, thus we set $\psi_L$ to be the linear activation function. Alternatively, the sigmoid function can be used  for binary classification  and the softmax function can be used for multi-class classification. The unknown parameters are weight matrices  $W^l$ and bias vectors $b^l$. The vectors $\mathbf{A^1},...,\mathbf{A^{L-1}}$ constitute the hidden layers. 

The network is trained by minimizing $\frac{1}{n} \sum_{i=1}^{n} l(Y_i - f(\mathbf{X_i}))$, where $l(u)$ is loss function. Different distributions of random error $\epsilon_i$ lead to different loss functions. If $\epsilon_i$ is assumed to be normal, then the likelihood is normally distributed and maximizing the log-likelihood is then equivalent to minimizing the squared loss function $l(u) = u^2$, and $f(\cdot)$ models the mean of the response values. If $\epsilon_i$ follows an asymmetric Laplace distribution, then maximizing the log-likelihood is then equivalent to minimizing the check loss function $l_{\gamma}(u) = u(\gamma - 1_{\{u<0\}})$, where $\gamma \in (0,1)$ refers to the quantile level, and $f(\cdot)$ models the $\gamma$ quantile of the responses. In this paper,  we consider these two assumptions about the random errors, and compare our 4N models with corresponding competitors. More details are given in Section 3. Note that even when $\epsilon_i$ is normally distributed, 4N is different than the NNGP model. 4N uses a multilayer perceptron for the link function $f(\cdot)$ to capture nonlinear dependence whereas NNGP restricts $f(\cdot)$ to be a linear combination of $Y_{\cal N_i}$. In other words, NNGP assumes a Gaussian joint density but 4N does not assume a parametric joint density.

\subsection{Feature Engineering}

4N models with different properties can be obtained by taking different summaries of the $m$ observations of the neighboring sets as features in $\mathbf{X}_i$.  The complete information in the neighboring set (assuming there are no covariates) are the $m$ observations $Y_{\cal N_i}$ and their $2m$ ($m$ latitudes and $m$ longitudes) locations $s_l$ for $l\in\cal N_i$.  We assume that the conditioning set is ordered by distance to location $s_i$, so that the location closest to $s_i$ appears first in $\cal N_i$ and the location farthest to $s_i$ appears last.  This ordering encourages the entries in $\mathbf{X}_i$ to have similar interpretation across observations.

A key feature that we use in $X_i$ is the Kriging prediction. The Kriging prediction is based on the model $Y_i = \mu+ W_i + \epsilon_i$,
where $\mu$ is the mean, $W_i$ is a mean-zero Gaussian process with correlation function $\phi(\cdot)$ and
 $\epsilon_i \sim N(0,\tau^2)$.   The Kriging prediction at a new location $s_i$ is
\[\hat{Y}_i = \hat{\mu} + \mathbf{C_{s_i}} \mathbf{C_{\cal N_i}^{-1}}(\mathbf{Y_{\cal N_i}} - \hat{\mu}), \]
where $\mathbf{C_{s_i}}$ is the $1 \times m$ covariance matrix (vector) between $W_i$ and $W_{\cal N_i}$, and $\mathbf{C_{\cal N_i}}$ is the $m \times m$ covariance matrix of $W_{\cal N_i}$. Since $\hat{Y}_i$ is a function only of information in the neighboring set, using it as a feature fit in the 4N framework.

Although the user is free to construct any features that are thought to be useful, we consider the following three sets of features:
\begin{itemize}
\item {\bf Kriging only}: In this case, $p=1$ and the feature is the Kriging prediction of $Y_i$ given $Y_{\cal N_i}$, denoted $X_{i1}={\hat Y}_i$. We use the exponential correlation function $cov(W_i, W_j) = \sigma^2\exp(-||\mathbf{s_i} - \mathbf{s_j}|| / \rho)$, where $\sigma^2$ is the variance, $\rho$ is the range parameter controlling spatial dependence and $||\cdot||$ is the Euclidean distance. Other correlation functions such as Matern and double exponential functions can also be used.   We estimate the spatial covariance parameters $\rho, \sigma^2$ and $\tau^2$ using Vecchia's approximation \citep{vecchia1988} and Permutation and Grouping Methods \citep{guinness2018}. We then derive the Kriging prediction using the equation above. This gives a stationary model but with potential non-linearity between the Kriging prediction and expected response. Since Kriging gives the best predictions under normality,  when data are generated from a Gaussian process and sample size is relatively small, 4N with Kriging as feature should give predictions close to ordinary Kriging.

\item {\bf Nonparametric features}: In this case, $p=3m+2$ and the features are $s_i$, $s_l-s_i$ ($m$ differences in latitudes and $m$ differences in longitudes) for $l\in\cal N_i$ and $Y_{\cal N_i}$. Including the differences $s_l-s_i$ allows for dependence as a function of distance, while including $s_i$ allows for nonstationarity, that is, different response surfaces in different locations.  The Kriging prediction ${\hat Y}_i$ is a function of these $p$ features and thus the model includes the parametric special case.  However, by including all $3m$ variables from the neighboring sites without any predefined structure we allow the neural network model to estimate the prediction equation nonparametrically. When data are not normally distributed, nonparametric features including neighboring information are anticipated to play a key role in prediction. Even if data are normally distributed, 4N with nonparametric features should approximate Kriging predictions as sample size increases.

\item {\bf Kriging + nonparametric (NP) features}: In this case, $p=3m+3$ as we add the Kriging prediction ${\hat Y}_i$ to the $3m+2$ features in the nonparametric model. We anticipate that the combination of the two sets of features above can allow 4N to have their respective advantages.  That is, if data are normal, including the Kriging feature should make the method competitive with Kriging, and if the true process is non-Gaussian and the sample size is large then the model should be able to learn complicated non-linear relationships and improve prediction.
\end{itemize}

\subsection{Computational Details}
We use the R package "GpGp" (\url{https://cran.r-project.org/web/packages/GpGp/GpGp.pdf}) to get estimates of the spatial parameters and subsequent Kriging prediction $\hat{Y}_i$ (Section 2.3). To implement 4N we use the deep learning package "keras" with GPU acceleration (Nvidia GeForce GTX 980Ti) on an 8-core i7-6700k machine with 16GB of RAM. We use Relu activation function for the hidden layers and linear activation for the output layers. For simulated and real data we set number of epochs to be 50 and impose early stopping and dropout (ranging from 0.1 to 0.5) to avoid overfitting. Early stopping is a technique to stop the training process early if the value of loss function does not decrease too much for several epochs. Dropout randomly makes nodes in the neural network dropped out by setting them to zero, which encourages the network to rely on other features that act as signals. That, in turn, creates more generalizable representations of the data.  We tune other hyperparameters using five-fold cross validation. For the simulation study, we tune the parameters with one dataset and use these pre-set tuning parameters for the other datasets. The parameters we tune are the learning rate, mini-batch size, number of layers and hidden units per layer. Learning rate is a hyperparameter that controls how much we are adjusting the weights of our network with respect to the loss gradient. We try different  values for the learning rate in the range of 0.1 to 1e-6. A mini-batch refers to the number of examples used at a time, when computing gradients and parameter updates.  Mini-batch size in the range of 16-128 are tried in our simulations and real data. We keep our number of layers to no more than three. As suggested in \cite{liang2018bayesian}, a
network with one hidden layer is usually large enough to approximate the system. We try a range of 100 to 500 hidden units for the first hidden layer and make number of hidden units gradually decrease for each layer. In general, number of hidden layers and hidden units per layer have less impact for model performance than learning rate, mini-batch size and regularization such as dropout and early stopping. All the parameters are trained using ADAM algorithm \citep{kingma2014adam}.

\section{Simulation Study}
We conduct simulations in a variety of settings to evaluate the performance of the 4N model with the three different sets of features. Here we focus on the 4N performance with respect to mean and quantile prediction by using squared loss function and check loss function, respectively. We give the details of the four simulation cases below.  We do not include any covariates to focus on modeling spatial dependence. Figure 1 plots a realization for each simulation case.
\begin{itemize}
\item {\bf Gaussian Process}: We generate data from the GP with mean $\mu=0$ and  exponential correlation function with variance $\sigma^2 = 5$, nugget variance $\tau^2 = 1$ and range parameter $\rho = 0.16$. 
\item {\bf Cubic and exponential transformation}: After generating data from the GP, we take $Y_i^3 / 100 + \exp(Y_i/5)/10$ as new response value. All the parameter settings for generating Gaussian process data are the same as above. 
\item {\bf Max stable process}: The max stable process is used for modeling spatial extremes. We generate data from the max stable process (in particular, the Schlather process of \cite{schlather2002})  using R package "spatialExtremes". The marginal distribution is the Frechet distribution with location parameter 1, scale parameter 2, shape parameter 0.3, and range parameter of the exponential correlation function 0.5. 
\item {\bf Potts model}: The Potts model \citep{potts1952} randomly assigns observations to clusters. Let $g_i \in \{1,...,G\}$ be the cluster label for the $i^{th}$ observation.  The Potts model can be defined by the distribution of $g_i$ given its neighboring set $\cal N_i$ as $\mbox{Prob}(g_i=k|g_l \mbox{ for }l\in{\cal N}_i) \propto \exp( \beta\sum_{l \in \cal N_i} \mathbf{1}\{g_l = k\})$, which encourages spatial clustering of the labels if $\beta>0$.   We simulate Potts model using Markov chain Monte Carlo algorithm (R package "potts"). In the simulation,  we generate $G = 8$ blocks and set $\beta=\log (1 + \sqrt{8})$. Given the cluster labels, the responses are generated as $Y_i|g_i \sim N(g_i^2+5g_i, \sqrt g_i)$. 
\end{itemize}

For each simulation setting data are generated on the unit square with the sample size $n$ set to be either 1,000 or 10,000. For each model setting and sample size, we generate 100 datasets. The mean predicted mean squared error, check loss and the corresponding standard error based on the 100 replications are shown in Tables 1 - 3. We compare the results from the three 4N models with those from NNGP and ALP. 
We implement Nearest-Neighbor Gaussian Process using R package "spNNGP" (\url{https://cran.r-project.org/web/packages/spNNGP/spNNGP.pdf}) for both parameter estimates and predictions. For Asymmetric Laplace Process, we use pacakage "baquantreg" (\url{https://github.com/brsantos/baquantreg/}) to get the parameter estimates and write our own codes to get predictions.

\begin{figure}[t]
\begin{subfigure}{.5\textwidth}
  \centering
  \includegraphics[width=.9\linewidth]{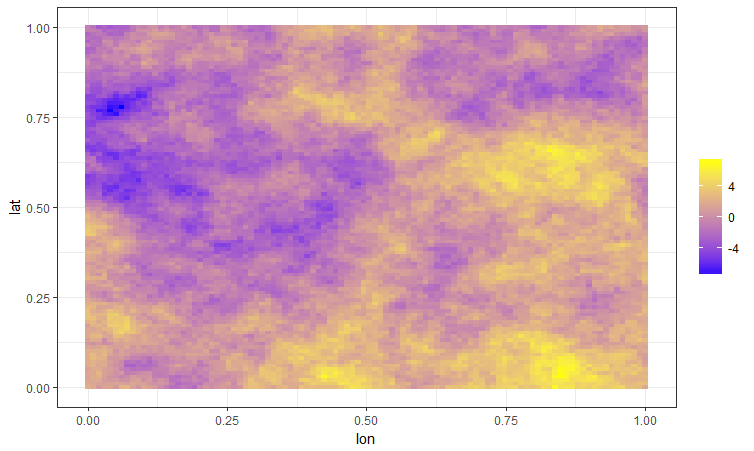}
  \caption{$Y_i$ generated from a Gaussian process}
  \label{fig:sfig1(a)}
\end{subfigure}%
\begin{subfigure}{.5\textwidth}
  \centering
  \includegraphics[width=.9\linewidth]{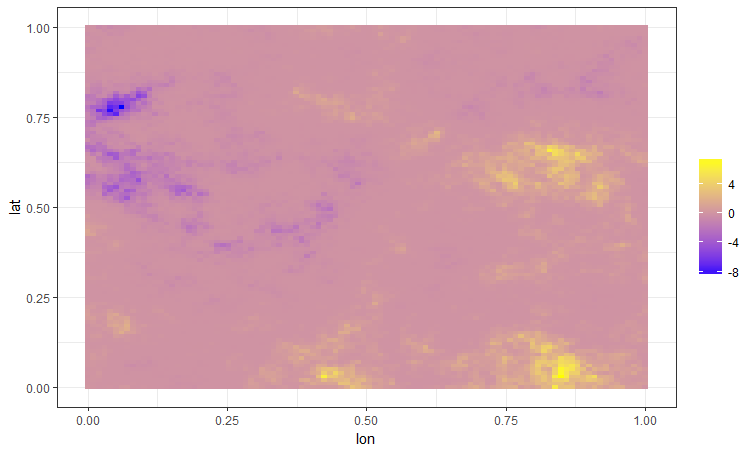}
  \caption{Exponential and cubic transformation of $Y_i$ generated from a Gaussian process}
  \label{fig:sfig1(b)}
\end{subfigure}
\begin{subfigure}{.5\textwidth}
  \centering
  \includegraphics[width=.9\linewidth]{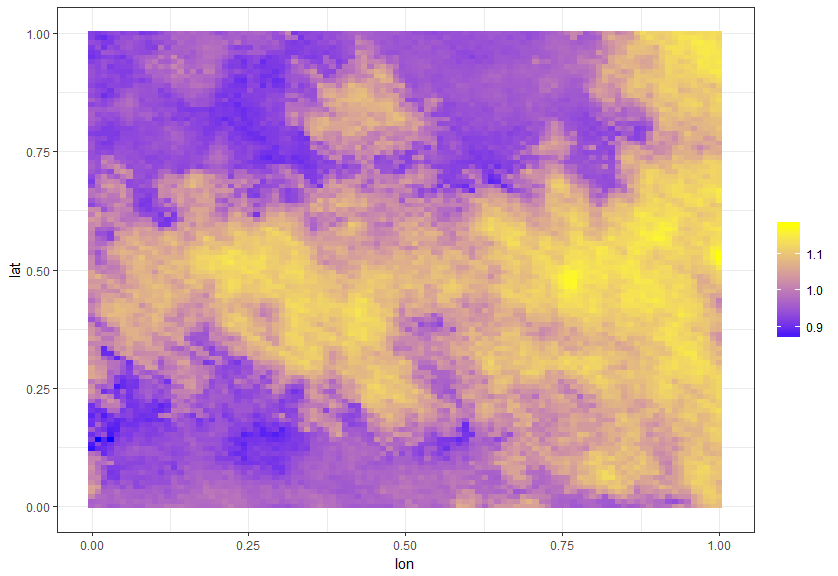}
  \caption{$Y_i$ generated from a max stable process}
  \label{fig:sfig1(c)}
\end{subfigure}%
\begin{subfigure}{.5\textwidth}
  \centering
  \includegraphics[width=.9\linewidth]{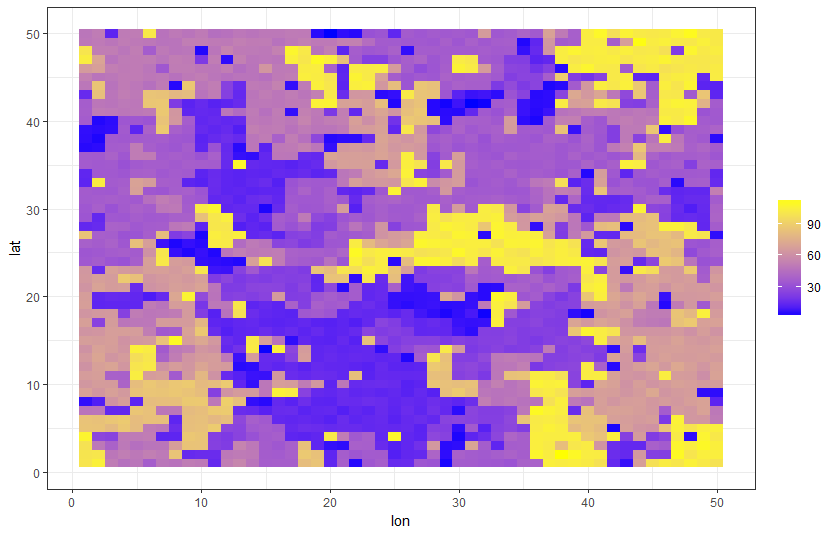}
  \caption{$Y_i$ generated from a Potts model}
  \label{fig:sfig1(d)}
\end{subfigure}
\caption{ Data ($\mathbf{Y}$) generated from different simulation settings.}
\label{fig:fig1}
\end{figure}

\subsection{Mean Prediction}

We first compare our 4N models with NNGP with respect to prediction mean squared error:
\[MSE = \frac{1}{k} \sum_{i=1}^k (Y_i - \hat{Y}_i)^2, \]
where $k$ is number of locations in testing dataset and $\hat{Y}_i$ is the predicted value at location $s_i$. Table 1 shows the results. For all methods we use $m=10$. When the data are generated from the GP, NNGP always has the best prediction results. The prediction MSE is at least $4.5\%$ smaller than the 4N model. Among the three 4N models, the Kriging only model gives the best results and the nonparametric features is the worst, as expected. However, when we take cubic and exponential transformation,  most of our 4N models have smaller MSE than NNGP.  For both $n=1,000$ and $n=10,000$ cases, 4N with only the Kriging prediction as a feature gives the best results, and MSE is $23\%$ and $28\%$ smaller than NNGP. This is expected because the data are generated from a transformation of GP and so the Kriging feature captures most of the spatial dependence.  For the max stable process and Potts models our 4N model with Kriging + nonparametric features gives the best prediction results. These processes are unrelated to a GP so the flexibility of including the nonparametric features improves prediction. 

Figure 2 further illustrates the role of sample size in the comparison between methods. We generate 50 replications of GP data using same parameters as above, but let $n=50,000$, $100,000$ and $500,000$. As sample size increases, the relative MSE of 4N over NNGP converges to one. For other models where 4N outperforms NNGP, such as the max stable process, the ratio is smaller than one and decreases as sample size increases.

We conclude from the simulation results that when data are normally distributed, NNGP gives better predictions than our 4N models, but the 4N models with the Kriging feature remain competitive. In both cases, the relative performance of 4N to NNGP improves as the sample size increases. However, for cases where data are not normally distributed, 4N can outperform NNGP.

\begin{table}[t]
\small
\centering
\caption{Mean squared error comparison between 4N models and NNGP (standard error shown in parenthesis) for simulated data.\label{tab1}}
\begin{tabular}{cccccc}
\toprule
\textbf{n}&\textbf{Method}&\textbf{GP}&\textbf{Transformed GP}&\textbf{Max Stable}&\textbf{Potts} \\
\midrule
1,000&4N(Kriging only)&2.11(0.18)&\textbf{0.71(0.06)}&4.85(0.11)&5.62(0.33)\\
		&4N(Nonparametric)&2.56(0.58)&0.89(0.08)&4.72(0.17)&5.43(0.71)\\
		&4N(Kriging+NP)&2.33(0.29)&0.72(0.07)&\textbf{4.64(0.18)}&\textbf{5.13(0.34)}\\
		&NNGP&\textbf{1.99(0.14)}&0.88(0.05)&4.81(0.19)&5.96(1.45)\\
		\midrule
			10,000&4N(Kriging only)&2.06(0.13)&\textbf{0.70(0.06)}&4.71(0.10)&5.55(0.23)\\
		&4N(Nonparametric)&2.48(0.48)&0.83(0.06)&4.57(0.16)&5.29(0.17)\\
		&4N(Kriging+NP)&2.23(0.23)&0.71(0.08)&\textbf{4.42(0.20)}&\textbf{5.12(0.19)}\\
		&NNGP&\textbf{1.97(0.13)}&0.85(0.09)&4.70(0.15)&5.92(0.18)\\
\bottomrule
\end{tabular}
\end{table}

\begin{figure}[t]
\centerline{\includegraphics[width=.5\linewidth]{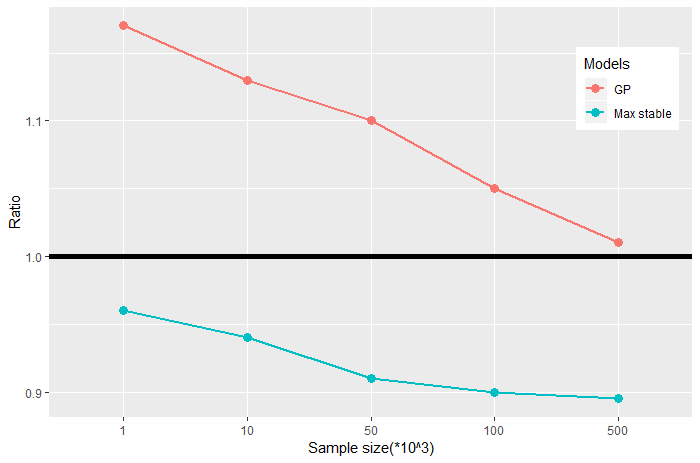}}
\caption{Relative prediction MSE of 4N (Kriging + Nonparametric) to NNGP as a function of sample size $n$, for data generated from GP and max stable models.\label{fig2}}
\end{figure}

\subsection{Quantile Prediction}
We also compare 4N models with the ALP model with respect to check loss:
\[l_{\gamma} = (\gamma - 1)\sum_{Y_i<\hat{Y_i}}(Y_i - \hat{Y_i}) + \gamma\sum_{Y_i\ge \hat{Y_i}}(Y_i - \hat{Y_i}), \]
where $\gamma$ is the quantile level of interest. We consider $\gamma\in\{0.25, 0.50, 0.75, 0.95\}$. Tables 2 ($n = 1,000$) and 3 ($n=10,000$) show the results. If data are generated from a Gaussian process, for $\gamma = 0.5$, ALP model has better prediction results than the 4N model with MSE $3\%$ to $8\%$ smaller than those from 4N models. However, for all other quantile levels, most of the 4N models are superior to the ALP model, with Kriging predictions + nonparametric features giving the best MSE, which is $5\%$ to $12\%$ smaller than the ALP model. Similar conclusions can be drawn for the max stable process except that when $\gamma = 0.95$, 4N with nonparametric features has slightly smaller MSE than 4N with Kriging predictions + nonparametric features ($1\%$ difference). For data generated as exponential and cubic transformations of Gaussian data, 4N with Kriging predictions + nonparametric features almost always have the best prediction results, except for $\gamma = 0.5$ where 4N with only Kriging prediction as feature has smaller MSE. Finally, for Potts model 4N models always outperform ALP model. 

\begin{table}[t]%
\small
\centering
\caption{Check loss with different quantiles comparison between 4N models and ALP model, with $n=1,000$ observations (standard error shown in parenthesis)  for simulated data.\label{tab2}}%
\begin{tabular}{ccccccc}
\toprule
\textbf{Quantile Level}&\textbf{Method}&\textbf{GP}&\textbf{Transformed GP}&\textbf{Max Stable}&\textbf{Potts} \\
\midrule
$0.95$&4N(Kriging only)&0.159(0.006)&0.477(0.029)&0.408(0.031)&0.273(0.045)\\
		&4N(Nonparametric)&0.150(0.005)&0.409(0.031)&\textbf{0.337(0.034)}&0.269(0.029)\\
		&4N(Kriging+NP)&\textbf{0.147(0.006)}&\textbf{0.370(0.027)}&0.341(0.029)&\textbf{0.267(0.023)}\\
		&ALP&0.154(0.005)&0.487(0.043)&0.381(0.039)&0.276(0.034)\\
		\midrule
		$0.75$&QuadN(Kriging only)&0.475(0.015)&1.121(0.054)&1.031(0.041)&0.695(0.025)\\
		&4N(Nonparametric)&0.501(0.026)&1.134(0.067)&1.034(0.034)&0.705(0.013)\\
		&4N(Kriging+NP)&\textbf{0.450(0.013)}&\textbf{1.083(0.035)}&\textbf{1.013(0.029)}&\textbf{0.680(0.028)}\\
		&ALP&0.506(0.014)&1.131(0.047)&1.118(0.032)&0.724(0.024)\\
		\midrule
		$0.5$&QuadN(Kriging only)&0.570(0.017)&\textbf{1.713(0.021)}&1.237(0.024)&0.770(0.031)\\
		&4N(Nonparametric)&0.593(0.026)&1.819(0.013)&1.264(0.033)&\textbf{0.705(0.019)}\\
		&4N(Kriging+NP)&0.566(0.015)&1.772(0.033)&1.226(0.012)&0.765(0.026)\\
		&ALP&\textbf{0.550(0.024)}&1.798(0.024)&\textbf{1.213(0.014)}&0.835(0.034)\\
		\midrule
		$0.25$&QuadN(Kriging only)&0.451(0.018)&0.928(0.029)&1.131(0.040)&0.671(0.035)\\
		&4N(Nonparametric)&0.472(0.058)&0.953(0.031)&1.127(0.040)&0.683(0.017)\\
		&4N(Kriging+NP)&\textbf{0.445(0.029)}&\textbf{0.902(0.027)}&\textbf{1.057(0.047)}&\textbf{0.647(0.022)}\\
		&ALP&0.466(0.014)&0.960(0.035)&1.100(0.019)&0.713(0.015)\\
\bottomrule
\end{tabular}
\end{table}

\begin{table}[t]%
\small
\centering
\caption{Check loss with different quantiles comparison between 4N models and ALP model, with $n=10,000$ observations (standard error shown in parenthesis)  for simulated data.\label{tab3}}%
\begin{tabular}{ccccccc}
\toprule
\textbf{Quantile Level}&\textbf{Method}&\textbf{GP}&\textbf{Transformed GP}&\textbf{Max Stable}&\textbf{Potts} \\
\midrule
$0.95$&4N(Kriging only)&0.154(0.007)&0.472(0.028)&0.409(0.034)&0.271(0.043)\\
		&4N(Nonparametric)&0.147(0.008)&0.407(0.030)&\textbf{0.335(0.024)}&0.264(0.023)\\
		&4N(Kriging+NP)&\textbf{0.143(0.009)}&\textbf{0.371(0.028)}&0.340(0.019)&\textbf{0.263(0.021)}\\
		&ALP&0.152(0.004)&0.480(0.047)&0.383(0.034)&0.277(0.038)\\
		\midrule
		$0.75$&QuadN(Kriging only)&0.471(0.012)&1.122(0.051)&1.029(0.044)&0.690(0.022)\\
		&4N(Nonparametric)&0.475(0.029)&1.110(0.063)&1.031(0.033)&0.680(0.019)\\
		&4N(Kriging+NP)&\textbf{0.451(0.015)}&\textbf{1.080(0.031)}&\textbf{1.011(0.022)}&\textbf{0.676(0.029)}\\
		&ALP&0.501(0.013)&1.130(0.044)&1.112(0.031)&0.720(0.022)\\
		\midrule
		$0.5$&QuadN(Kriging only)&0.567(0.017)&\textbf{1.708(0.022)}&1.233(0.022)&0.751(0.030)\\
		&4N(Nonparametric)&0.570(0.023)&1.810(0.018)&1.260(0.032)&\textbf{0.700(0.011)}\\
		&4N(Kriging+NP)&0.562(0.017)&1.773(0.032)&1.222(0.011)&0.723(0.029)\\
		&ALP&\textbf{0.547(0.021)}&1.792(0.023)&\textbf{1.210(0.012)}&0.732(0.032)\\
		\midrule
		$0.25$&QuadN(Kriging only)&0.444(0.016)&0.950(0.023)&1.132(0.042)&0.672(0.032)\\
		&4N(Nonparametric)&0.450(0.054)&0.923(0.032)&1.123(0.044)&0.650(0.015)\\
		&4N(Kriging+NP)&\textbf{0.443(0.023)}&\textbf{0.901(0.023)}&1.054(0.043)&\textbf{0.642(0.021)}\\
		&ALP&0.463(0.014)&0.961(0.033)&\textbf{1.004(0.013)}&0.710(0.013)\\
\bottomrule
\end{tabular}
\end{table}

\subsection{Variable Importance}
In addition to prediction accuracy, we would like to understand the dependence structure estimated by the 4N model for each simulation scenario. To explore the dependence structure we compute the relative importance of each feature \cite[for a review of importance measures, see][]{gevrey2003review}. We use \cite{Garson:1991:INC:129449.129452}'s method to partition the connection weight $\mathbf{W^l}$ and bias $\mathbf{b^l}$ to determine the relative importance of the input variables. The method essentially involves partitioning the hidden-output connection weights of each hidden neuron into components associated with each input neuron. By construction, the sum of the relative importance over the $p$ features is one. 

For the simulation with $n=10,000$, we calculate the relative importance for each input variable using Kriging + nonparametric model and nonparametric model averaged over the 100 replications. Figure 3 shows the results. The importance of $s_l$ and $Y_l$ are combined for each neighbor, and the results are plotted so $l=1$ is the nearest neighbor and $l=10$ is the most distant neighbor. Figure 3 also includes the importance of the Kriging feature $\hat{Y_i}$ and the prediction location $s_i$ to measure the effect of nonstationarity. For the Kriging + nonparametric model, if the data are generated from a GP or max stable process, then the Kriging feature is the most important.  However, for the transformed GP or Potts model importance of Kriging feature decreases and is less than most of the neighboring information. For the other features, importance decreases from the closest neighbor to the farthest neighbor, which is shown more clearly in Figure 3(b) for the nonparametric model. For GP data the closest neighbor is more important than the others, consistent with Kriging prediction. Similar conclusion can be made for data from max stable process.  However, for the other scenarios the importance decreases more slowly so that distant neighbors are almost as important as the closer neighbors. Across all the models, the spatial location of the prediction site is the least important feature, which is as expected because the data are generated from stationary processes.

\begin{figure}[t]
\begin{subfigure}{.5\textwidth}
  \centering
  \includegraphics[width=.9\linewidth]{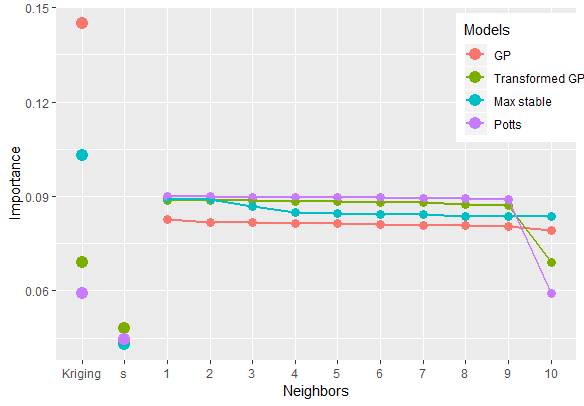}
  \caption{Kriging + nonparametric}
  \label{fig:sfig3(a)}
\end{subfigure}%
\begin{subfigure}{.5\textwidth}
  \centering
  \includegraphics[width=.9\linewidth]{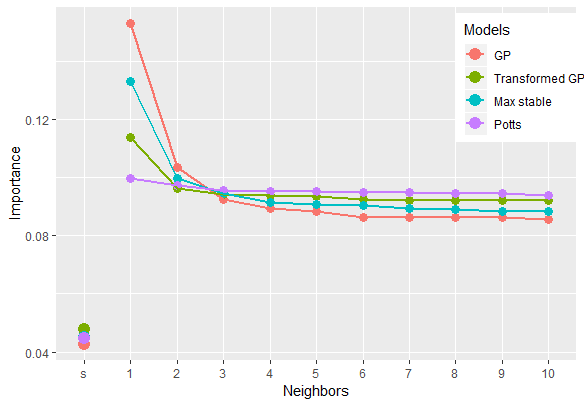}
  \caption{Nonparametric}
  \label{fig:sfig3(b)}
\end{subfigure}
\caption{Relative variable importance (averaged over simulated datasets) for the Kriging feature, spatial location of the prediction site, and the total importance of the three features (latitude and longitude, difference to the prediction location, and response) for each neighbor, plotted so that $l=1$ is the closest neighbor and $l=10$ is the most distant.}
\label{fig:fig3}
\end{figure}

\section{Canopy height data analysis}\label{sec5}
\begin{figure}[t]
\begin{subfigure}{.5\textwidth}
  \centering
  \includegraphics[width=.9\linewidth]{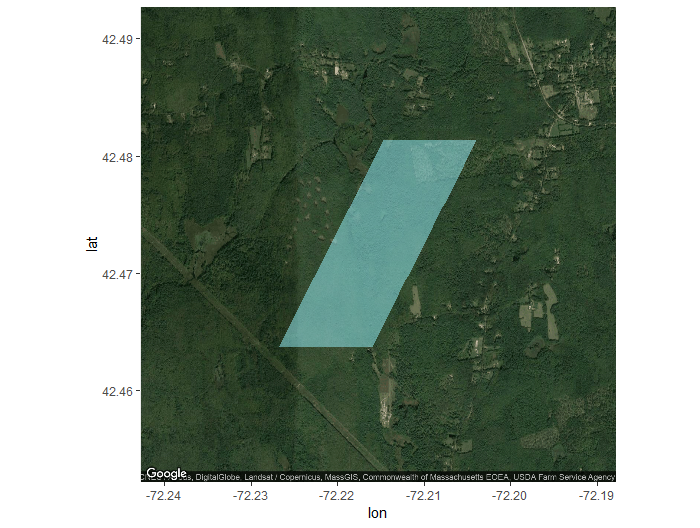}
  \caption{}
  \label{fig:sfig4(a)}
\end{subfigure}%
\begin{subfigure}{.5\textwidth}
  \centering
  \includegraphics[width=.9\linewidth]{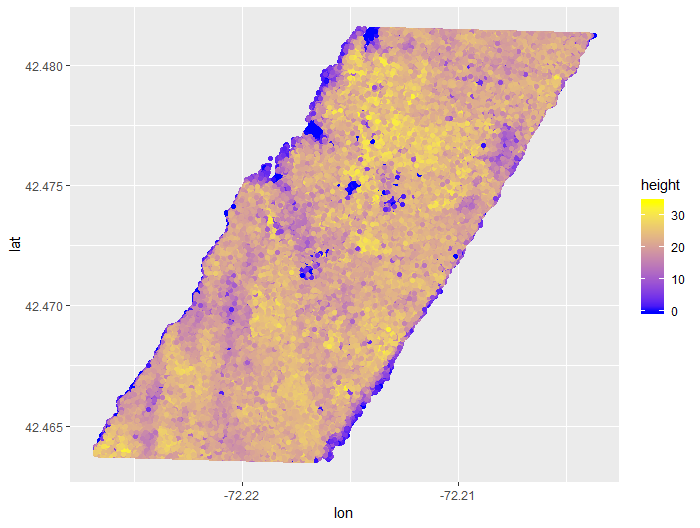}
  \caption{}
  \label{fig:sfig4(b)}
\end{subfigure}
\caption{(a) Satellite image from Google map, with the blue diamond showing the area where the data are collected and (b) canopy height (meters) in the study area.}
\label{fig:fig4}
\end{figure}

In an effort to quantify forest carbon baseline and change, researchers need complete maps of forest biomass. These maps are used in forest management and global carbon monitoring and modeling efforts. Globally, canopy height represents a substantial source and sink in the global carbon cycle. Canopy Height Model (CHM) from NASA Goddard's LiDAR  Hyperspectral and Thermal (G-LiHT) \citep{cook2013nasa} Airborne Imager over a subset of Harvard Forest Simes Tract, MA, was collected in Summer 2012. G-LiHT is a portable multi-sensor system that is used to collect fine spatial-scale image data of canopy structure, optical properties, and surface temperatures. Data we analyze here are downloaded from \url{https://gliht.gsfc.nasa.gov} and contain 1,723,137 observations. Locations are in UTM Zone 18 in the raw data, we convert them into standard longitude and latitude convention in the analysis. Figure \ref{fig:fig4} plots the data. 

The same models, fitting procedures and performance metrics are applied to these data as in the simulation study.
We subset 500,000 observations of the original dataset as training data and take the rest as testing data. The training dataset is further split with $90\%$ for training and $10\%$ for validation to tune the model. 
Table 4 shows the results. The NNGP method has the smallest MSE for estimating the mean, but all methods are similar.  However, the 4N models outperform ALP model for quantile prediction in all cases. For $\gamma = 0.95$, the Kriging-only 4N model is the best with check loss 10\% less than ALP model. For other $\gamma$ values, Kriging + nonparametric model gives the best results, which are 10\% to 14\% better than ALP.

We also conduct a variable importance analysis for the CHM data (Figure 5).  For all objective functions the Kriging feature is the most important, especially for MSE. The importance of neighboring information decreases with distance at a similar rate for all five objective functions.  The spatial-location features account for approximately 5\% of the total importance, suggesting nonstationarity is not strong for this analysis.

\begin{table}[t]%
\small
\centering
\caption{Prediction performance for 4N models, NNGP and ALP for the canopy height data.\label{tab4}}%
\begin{tabular}{lccccc}
\toprule
\textbf{Loss function}&\textbf{4N(Kriging only)}&\textbf{4N(Nonparametric)}&\textbf{4N(Kriging+NP)}&\textbf{NNGP}&\textbf{ALP} \\
\midrule
MSE&1.435&1.436&1.433&\textbf{1.430}&NA\\
Check loss ($\gamma$=0.25)&1.283&1.184&\textbf{1.157}&NA&1.279\\
Check loss ($\gamma$=0.5)&1.350&1.364&\textbf{1.301}&NA&1.489\\
Check loss ($\gamma$=0.75)&1.112&1.194&\textbf{1.034}&NA&1.156\\
Check loss ($\gamma$=0.95)&\textbf{0.360}&0.413&0.386&NA&0.401\\
\bottomrule
\end{tabular}
\end{table}

\begin{figure}[t]
\begin{subfigure}{.5\textwidth}
  \centering
  \includegraphics[width=.9\linewidth]{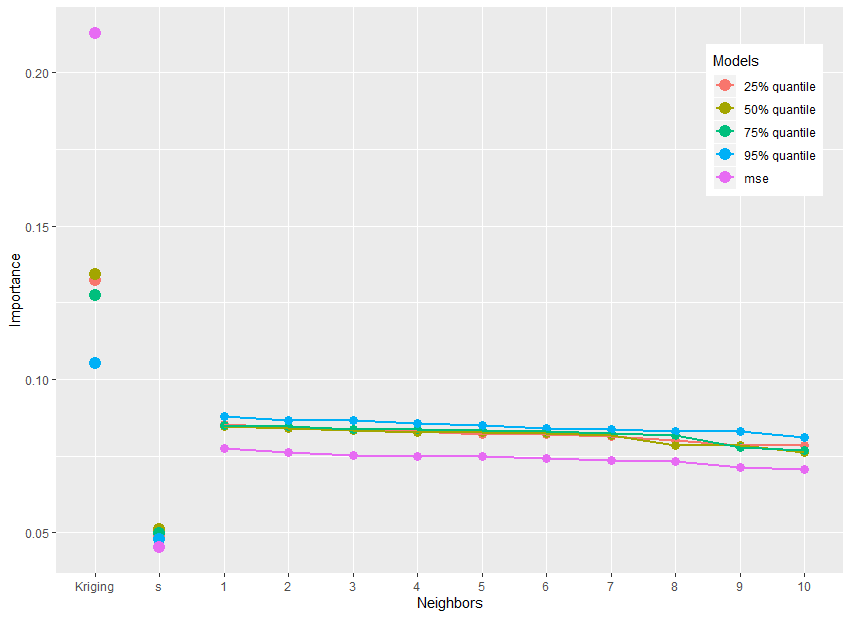}
  \caption{}
  \label{fig:sfig5(a)}
\end{subfigure}%
\begin{subfigure}{.5\textwidth}
  \centering
  \includegraphics[width=.9\linewidth]{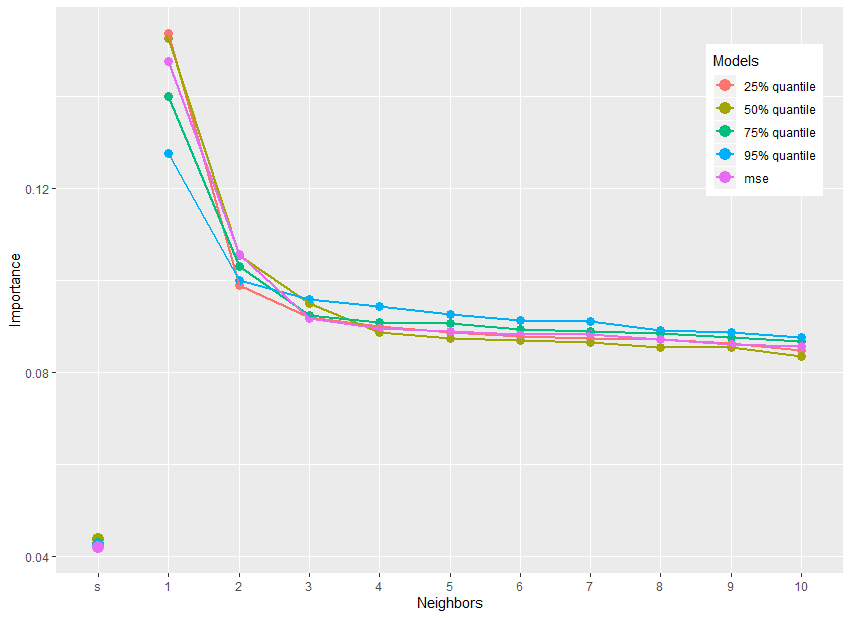}
  \caption{}
  \label{fig:sfig5(b)}
\end{subfigure}
\caption{Relative Variable Importance for CHM Data}
\label{fig:fig5}
\end{figure}

\section{Conclusions}\label{sec6}
In this paper, we apply deep learning to geostatistical prediction. We formulate a flexible model that is a valid stochastic process and propose a series of new ways to construct features based on neighboring information. Our  method outperforms state-of-art geostatistical methods for non-normal data. We also provide practical guidance on using MLP for modeling point-referenced geostatistical data that is fast and easy to implement. The 4N framework can be easily extended to other prediction tasks, such as classification, by modifying the activation and loss functions.  In addition, the implementation of 4N models can utilize powerful GPU computing for further acceleration. 

In geostatistics it is often desirable to attach a measure of uncertainty to spatial predictions and uncertainty quantification is known to be a challenge for deep learning methods. A Bayesian implementation naturally provides measures of uncertainty, but at additional computational costs.  Neural networks are related to Gaussian processes \citep{neal1996priors, lee2018deep} and this relationship can be used to provide uncertainty quantification \citep{graves2011practical, kingma2015variational, blundell2015weight}. \cite{gal2016dropout} prove dropout training in deep neural networks can be regarded as approximate Bayesian inference in deep Gaussian processes. One possibility that arises from our paper is to use quantile predictions resulting from check-loss optimization (e.g., Table 2) to form 95\% prediction intervals.  Applying this method in the simulation study gave coverage 98\% for the GP, 94\% for the transformed GP, 94\% for the max stable process and 87\% for the Potts model. For the CHM data, the 95\% prediction interval coverage is exactly 95\%.  Therefore this approach warrants further study.

\begin{singlespace}
	\bibliographystyle{rss}
	\bibliography{GrainBoundary}
\end{singlespace}

\end{document}